\documentclass{article}

\PassOptionsToPackage{numbers, compress}{natbib}
\usepackage[nonatbib,final]{nips_2016}

\usepackage{nips_2016}

\usepackage{color,soul}
\usepackage{subfigure} 
\usepackage{mathtools}
\DeclarePairedDelimiter{\ceil}{\lceil}{\rceil}

\usepackage[utf8]{inputenc} 
\usepackage[T1]{fontenc}    
\usepackage{hyperref}       
\usepackage{url}            
\usepackage{booktabs}       
\usepackage{amsfonts}       
\usepackage{nicefrac}       
\usepackage{microtype}      
\usepackage[pdftex]{graphicx}
\usepackage{amsmath,amssymb}

\usepackage{bm}

\usepackage{color,soul}

\newcommand{\eob}{\textless$e$\textgreater}

\title{A Neural Transducer}

\author{
  Navdeep Jaitly \\
  Google Brain \\
  \texttt{ndjaitly@google.com} \\
  \And
  David Sussillo \\
  Google Brain \\
  \texttt{sussillo@google.com} \\
  \And
  Quoc V. Le \\
  Google Brain \\
  \texttt{qvl@google.com} \\
  \And
  Oriol Vinyals \\
  Google DeepMind \\
  \texttt{vinyals@google.com} \\
  \And
  Ilya Sutskever \\
  Open AI\thanks{Work done at Google Brain} \\
  \texttt{ilyasu@openai.com} \\
  \And
  Samy Bengio \\
  Google Brain \\
  \texttt{bengio@google.com}
}

\begin{document}

\maketitle
\begin{abstract}
Sequence-to-sequence models have achieved impressive results on
various tasks.  However, they are unsuitable for tasks that require
incremental predictions to be made as more data arrives or tasks that
have long input sequences and output sequences. This is because they
generate an output sequence conditioned on an entire input
sequence. In this paper, we present a \emph{Neural Transducer} that
can make incremental predictions as more input arrives, without
redoing the entire computation.  Unlike sequence-to-sequence models,
the Neural Transducer computes the next-step distribution conditioned
on the {\it partially} observed input sequence and the partially
generated sequence. At each time step, the transducer can decide to emit
zero to many output symbols. The data can be processed using an encoder
and presented as input to the transducer. The discrete decision to emit
a symbol at every time step makes it difficult to learn with conventional
backpropagation. It is
however possible to train the transducer by using a dynamic programming
algorithm to generate target discrete decisions. Our experiments show
that the Neural
Transducer works well in settings where it is required to produce
output predictions as data come in. We also find that the Neural
Transducer performs well for long sequences even when attention
mechanisms are not used.

\end{abstract}

\section{Introduction}
The recently introduced sequence-to-sequence model has shown success
in many tasks that map sequences to sequences, e.g., translation,
speech recognition, image captioning and dialogue
modeling~\cite{sutskever-nips-2014,cho-emnlp-2014,bahdanau-iclr-2015,
  chorowski-nips-2015,chan2015listen,vinyals-arvix-2014,vinyals2015grammar,sordoni2015neural,vinyals2015neural}.
However, this method is unsuitable for tasks where it is important to
produce outputs as the input sequence arrives. Speech recognition is
an example of such an {\it online} task -- users prefer seeing an
ongoing transcription of speech over receiving it at the ``end'' of an
utterance.  Similarly, instant translation systems would be much more
effective if audio was translated online, rather than after entire
utterances. This limitation of the sequence-to-sequence model is due
to the fact that output predictions are conditioned on the entire input
sequence.

\begin{figure}[t]
\begin{center}
\subfigure[seq2seq]{
 \includegraphics[scale=0.22,clip=true,trim=7cm 9cm 4.5cm 8cm]{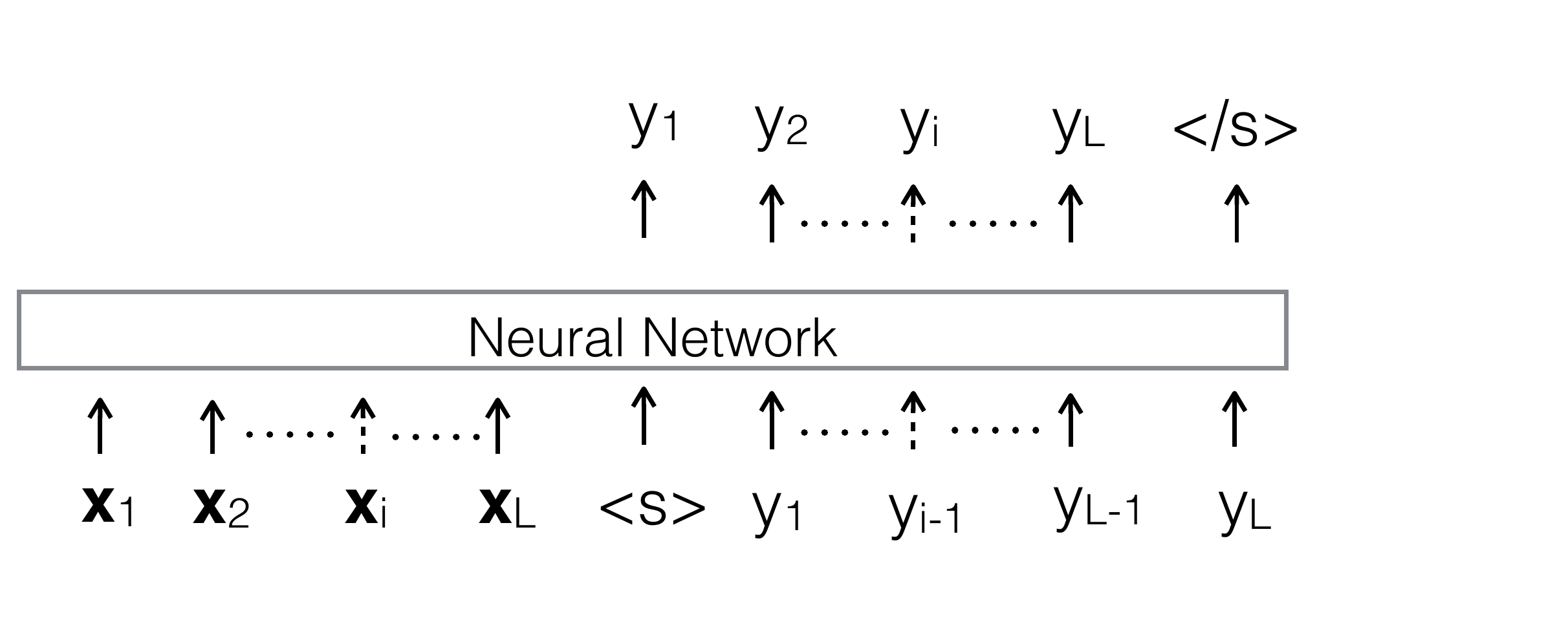}
}
\subfigure[Neural Transducer]{
 \includegraphics[width=0.4\textwidth,clip=true,trim=5.5cm 9cm 6.2cm 4cm]{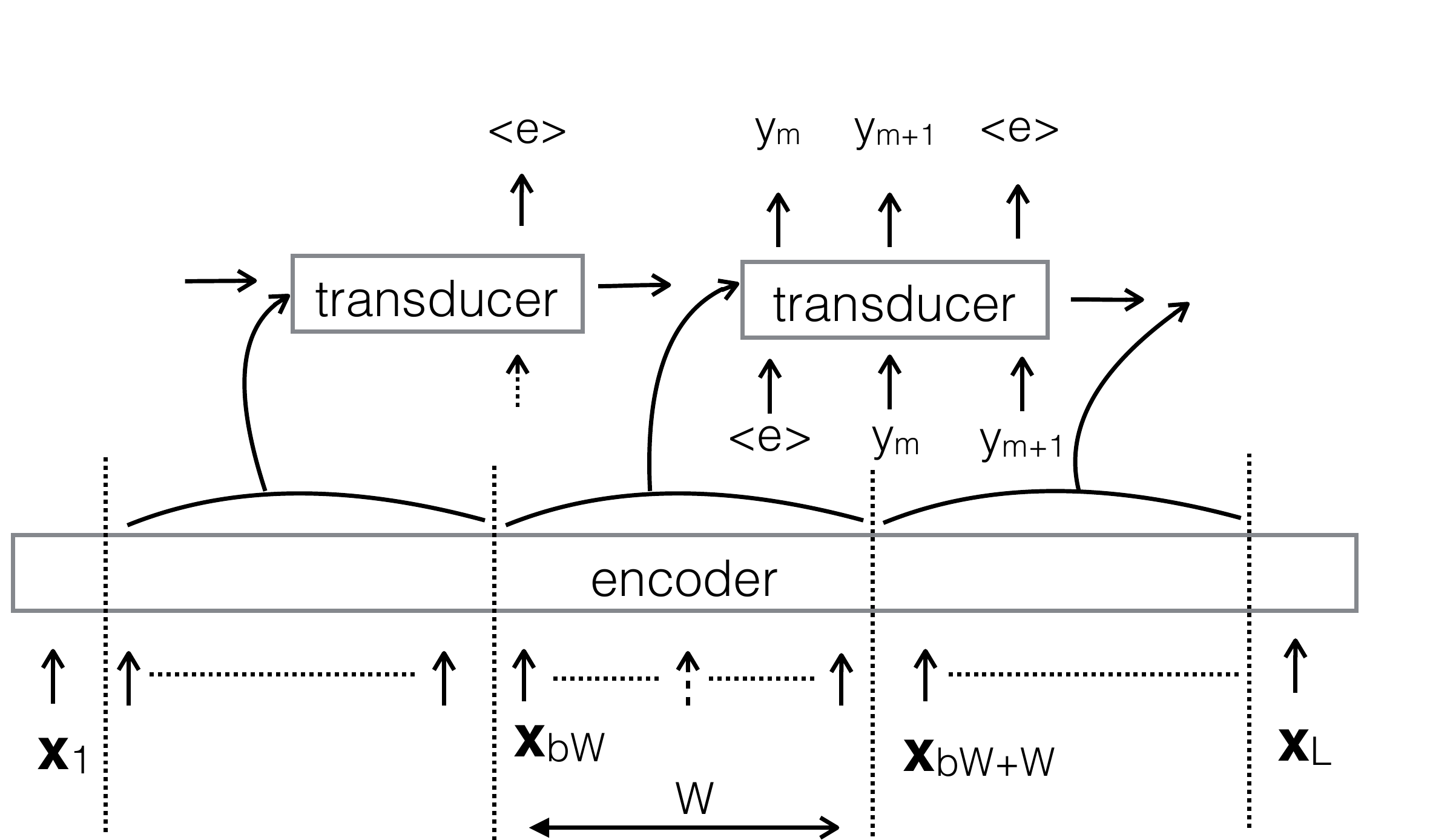}
}

\end{center}
\caption{High-level comparison of our method with sequence-to-sequence
  models. (a) Sequence-to-sequence model~\cite{sutskever-nips-2014}.
(b) The Neural Transducer (this paper) which emits output symbols as
  data come in (per block) and transfers the hidden state across
  blocks.
}
\label{fig:comparison}
\vspace{-0.2in}
\end{figure}

In this paper, we present a Neural Transducer, a more general class of
sequence-to-sequence learning models.  Neural Transducer can produce
chunks of outputs (possibly of zero length) as blocks of inputs arrive
- thus satisfying the condition of being ``online'' (see
Figure~\ref{fig:comparison}(b) for an overview).  The model generates
outputs for each block by using a transducer RNN that implements a
sequence-to-sequence model.  The inputs to the transducer RNN come
from two sources: the encoder RNN and its own recurrent state. In
other words, the transducer RNN generates local extensions to the
output sequence, conditioned on the features computed for the block by
an encoder RNN and the recurrent state of the transducer RNN at the
last step of the previous block.





During training, alignments of output symbols to the input sequence are
unavailable. One way of overcoming this limitation is to treat the alignment
as a latent variable and to marginalize over all possible values of this
alignment variable. Another approach is to generate alignments from a
different algorithm, and train our model to maximize the probability of these
alignments. Connectionist Temporal Classification (CTC)~\cite{graves2013speech}
follows the former strategy using a dynamic programming algorithm, that allows
for easy marginalization over the unary potentials produced by a recurrent
neural network (RNN). However, this is not possible in our model, since
the neural network makes next-step predictions that are conditioned not
just on the input data, but on the alignment, and the targets produced
until the current step. In this paper, we show how a dynamic programming
algorithm, can be used to compute "approximate" best alignments from this
model. We show that training our model on these alignments leads to
strong results.

On the TIMIT phoneme recognition task, a Neural Transducer (with 3 layered
unidirectional LSTM encoder and 3 layered unidirectional LSTM transducer) can achieve an
accuracy of 20.8\% phoneme error rate (PER) which is close to state-of-the-art for
unidirectional models. We show too that if good alignments are made
available (e.g, from a GMM-HMM system), the model can achieve 19.8\%
PER.

\section{Related Work}
In the past few years, many proposals have been made to add more power
or flexibility to neural networks, especially via the concept of
augmented
memory~\cite{graves2014neural,sukhbaatar2015end,zaremba2015reinforcement}
or augmented arithmetic
units~\cite{neelakantan2015neural,reed2015neural}. Our work is not
concerned with memory or arithmetic components but it allows more
flexibility in the model so that it can dynamically produce outputs as
data come in.

Our work is related to traditional structured prediction methods,
commonplace in speech recognition.  The work bears similarity to
HMM-DNN~\cite{hinton2012deep} and CTC~\cite{graves2013speech} systems. An
important aspect of these approaches is that the model makes
predictions at every input time step. A weakness of these models is
that they typically assume conditional independence between the
predictions at each output step.




Sequence-to-sequence models represent a breakthrough where no such
assumptions are made -- the output sequence is generated by next step
prediction, conditioning on the entire input sequence and the partial
output sequence generated so far~\cite{chorowski-nips-2014,chorowski-nips-2015,chan2015listen}.
Figure~\ref{fig:comparison}(a) shows the high-level
picture of this architecture. However, as can be seen from the figure,
these models have a limitation in that they have to wait until the end
of the speech utterance to start decoding. This property makes them
unattractive for real time speech recognition and online translation.
Bahdanau et. al.~\cite{BahdanauCSBB15} attempt to rectify
this for speech recognition by using a moving windowed attention, but they
do not provide a mechanism to address the situation that arises when no
output can be produced from the windowed segment of data.

Figure~\ref{fig:comparison}(b) shows the
difference between our method and sequence-to-sequence models.

A strongly related model is the \emph{sequence transducer}~\cite{graves-icml-2012,graves-icassp-2013}.
This model augments the CTC model by combining the transcription model with a
prediction model. The prediction model is akin to a language model and operates only
on the output tokens, as a next step prediction model. This gives the model more
expressiveness compared to CTC which makes independent predictions at every time step. However, unlike
the model presented in this paper, the two models in the sequence transducer operate independently --
the model does not provide a mechanism by which the prediction network features at one time step
would change the transcription network features in the future, and vice versa. Our model,
in effect both generalizes this model and the sequence to sequence model.

Our formulation requires inferring alignments during training. However,
our results indicate that this can be
done relatively fast, and with little loss of accuracy, even on a small
dataset where no effort was made at regularization. Further, if
alignments are given, as is easily done offline for various tasks,
the model is able to train relatively fast, without this inference step.

\section{Methods}
\begin{figure}[t]
\begin{center}
 \includegraphics[width=0.75\textwidth,trim=5cm 5cm 2cm 0mm]{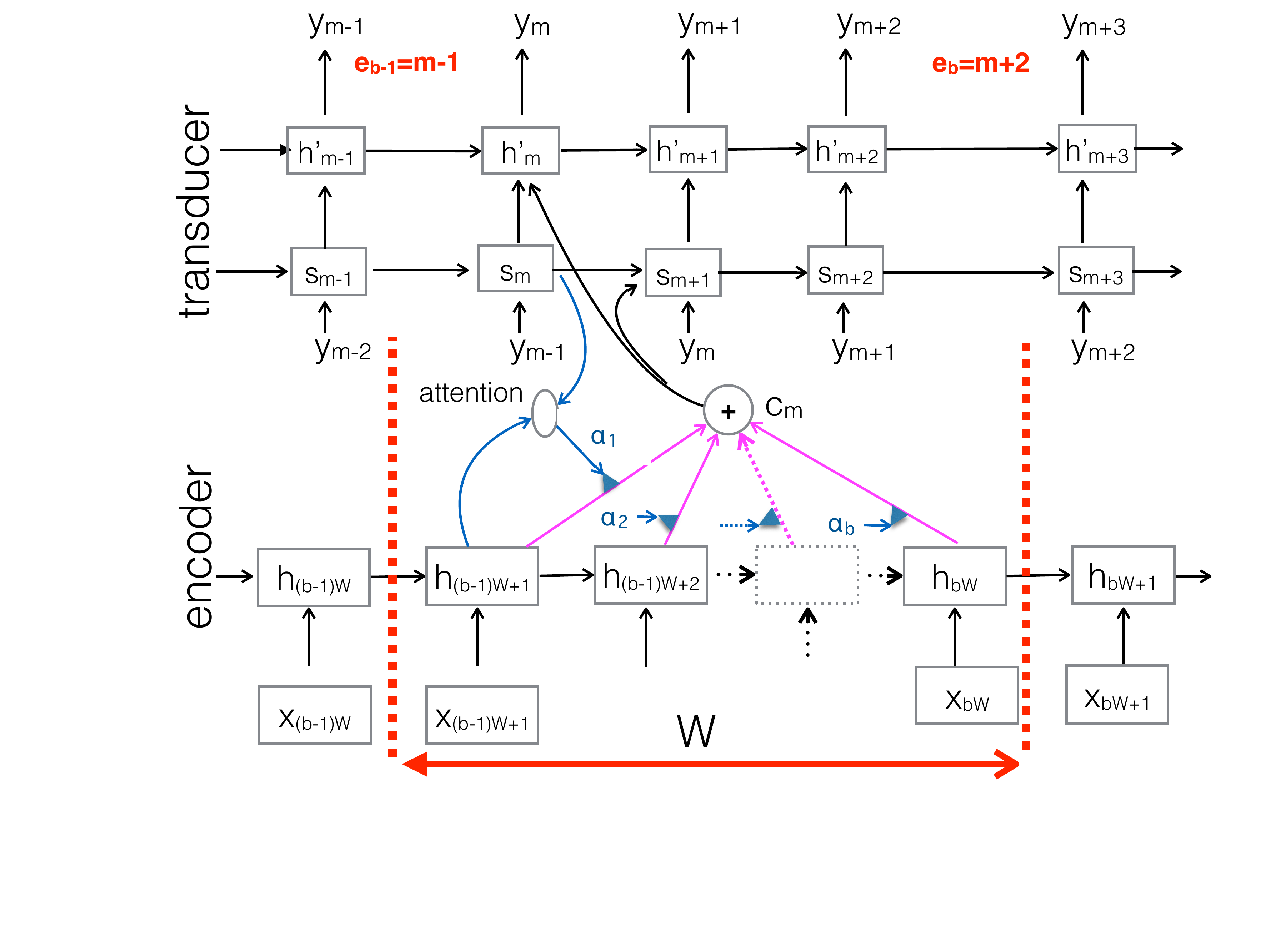}
\end{center}
\caption{\small An overview of the Neural Transducer architecture
  for speech. The input acoustic sequence is processed by the encoder
  to produce hidden state vectors ${\bf h}_i$ at each time step $i$,
  $i=1 \cdots L$.  The transducer receives a block of inputs at each step
  and produces up to $M$ output tokens using the sequence-to-sequence
  model over this input. The transducer maintains its state across the
  blocks through the use of recurrent connections to the previous output
  time steps. The figure above shows the transducer producing tokens
  for block $b$. The subsequence emitted in this block is $y_m y_{m+1} y_{m+2}$.
  \label{fig:overview}
  }
\end{figure}

In this section we describe the model in more detail. Please refer to
Figure~\ref{fig:overview} for an overview.

\subsection{Model}
Let ${\bf x}_{1 \cdots L}$ be the input data that is $L$ time steps
long, where ${\bf x}_i$ represents the features at input time step
$i$.  Let $W$ be the block size, i.e., the periodicity with which the
transducer emits output tokens, and $N=\ceil{\frac{L}{W}}$ be the
number of blocks.

Let $\tilde{y}_{1 \cdots S}$ be the target sequence, corresponding to the
input sequence. Further, let the transducer produce a sequence of
$k$ outputs, $\tilde{y}_{i \cdots \left(i+k\right)}$, where $0 \le k < M$, for
any input block.
Each such sequence is padded with the $\eob$ symbol, that is added to the
vocabulary.  It signifies that the transducer may proceed and consume data
from the next block.  When no symbols are produced for a block, this
symbol is akin to the blank symbol of CTC.

The sequence $\tilde{y}_{1 \cdots S}$ can be transduced from the input
from various alignments.  Let $\mathcal{Y}$ be the set of all alignments
of the output sequence $\tilde{y}_{1 \cdots S}$ to the input blocks.
Let $y_{1 \cdots \left(S+B)\right)} \in{\mathcal{Y}}$ be any
such alignment. Note that the length of $y$ is $B$ more than the length of
$\tilde{y}$, since there are $B$ end of block symbols, $\eob$, in $y$.
However, the number of sequences $y$ matching to $\tilde{y}$ is much
larger, corresponding to all possible alignments of $\tilde{y}$ to the blocks.
The block that element $y_i$ is aligned to can be inferred
simply by counting the number of $\eob$ symbols that came before index $i$.
Let, $e_b, b\in 1 \cdots N$ be the index of the last token in $y$ emitted in
the $b^{th}$ block. Note that $e_0 = 0$ and $e_N = \left(S+B\right)$.
Thus $y_{e_b}=$\eob \hspace{0.03in}for each block $b$.

In this section, we show how to compute $p\left(y_{1 \cdots \left(S+B)\right)} | {\bf x}_{1 \cdots L} \right)$.
Later, in section ~\ref{noalignment} we show how to compute, and maximize $p\left(\tilde{y}_{1 \cdots S} | {\bf x}_{1 \cdots L} \right)$.

We first compute the probability of l compute the probability of seeing output sequence $y_{1 \cdots e_b}$ by the end of
block $b$ as follows:
\begin{equation}
	p\left(y_{1 \cdots e_b} | {\bf x}_{1 \cdots bW}\right) = p\left(y_{1 \cdots e_1} | {\bf x}_{1 \cdots W} \right)
	\prod_{b'=2}^{b} p\left(y_{\left(e_{b'-1}+1\right) \cdots e_b'} | {\bf x}_{1 \cdots b'W}, y_{1 \cdots e_{b'-1}} \right)
\end{equation}
Each of the terms in this equation is itself computed by the chain rule
decomposition, i.e., for any block $b$,
\begin{equation}
\label{eqn:next_step}
	p\left(y_{\left(e_{b-1}+1\right) \cdots e_{b}} | {\bf x}_{1 \cdots bW}, y_{1 \cdots e_{b-1}} \right) =
\prod_{m=e_{\left(b-1\right)}+1}^{e_{b}}p\left(y_{m}| {\bf x}_{1 \cdots bW}, y_{1 \cdots \left(m-1\right)} \right)
\end{equation}

The next step probability terms, $p\left(y_m| {\bf x}_{1 \cdots bW},
y_{1 \cdots \left(m-1\right)} \right)$, in
Equation~\ref{eqn:next_step} are computed by the transducer using the
encoding of the input ${\bf x}_{1 \cdots bW}$ computed by the encoder,
and the label prefix $y_{1 \cdots \left(m-1\right)}$ that was input
into the transducer, at previous emission steps. We describe this in
more detail in the next subsection.

\subsection{Next Step Prediction}
We again refer the reader to Figure~\ref{fig:overview} for this
discussion. The example shows a transducer with two hidden layers,
with units ${\bf s}_m$ and ${\bf h'}_m$ at output step $m$. In the
figure, the next step prediction is shown for block $b$. For this
block, the index of the first output symbol is $m=e_{b-1}+1$, and the
index of the last output symbol is $m+2$ (i.e.  $e_b=m+2$).

The transducer computes the next step prediction, using parameters, ${\bf \theta}$, of the
neural network through the following sequence of steps:
\begin{eqnarray}
{\bf s}_m &=& f_{RNN}\left({\bf s}_{m-1}, \left[{\bf c}_{m-1}; y_{m-1}\right];{\bf \theta}\right)\\
{\bf c}_m &=& f_{context}\left({\bf s}_m, {\bf h}_{\left(\left(b-1\right)W+1\right) \cdots bW};{\bf \theta}\right)\\
{\bf h'}_m &=& f_{RNN}\left({\bf h'}_{m-1}, \left[{\bf c}_m; s_m\right];{\bf \theta}\right)\\
p\left(y_m | {\bf x}_{1 \cdots bW}, y_{1 \cdots \left(m-1\right)} \right) 
 &=& f_{softmax}\left(y_m ; {\bf h'}_m,{\bf \theta}\right) \label{f_softmax}
\label{eqn:transd}
\end{eqnarray}
where $f_{RNN}\left({\bf a_{m-1}},{\bf b_m};{\bf \theta}\right)$ is
the recurrent neural network function (such as an LSTM or a sigmoid or
tanh RNN) that computes the state vector ${\bf a}_m$ for a layer at a
step using the recurrent state vector ${\bf a}_{m-1}$ at the last time
step, and input ${\bf b}_m$ at the current time step;\footnote{Note
  that for LSTM, we would have to additionally factor in cell states
  from the previous states - we have ignored this in the notation for
  purpose of clarity. The exact details are easily worked out.}
$f_{softmax}\left(\cdot;{\bf a_m};{\bf \theta}\right)$ is the softmax
distribution computed by a softmax layer, with input vector ${\bf
  a_m}$; and $f_{context}\left({\bf s}_m, {\bf
  h}_{\left(\left(b-1\right)W+1\right) \cdots bW};{\bf \theta}\right)$
is the context function, that computes the input to the transducer at
output step $m$ from the state $s_m$ at the current time step, and the
features ${\bf h}_{\left(\left(b-1\right)W+1\right) \cdots bW}$ of the
encoder for the current input block, $b$. We experimented with
different ways of computing the context vector -- with and without an
attention mechanism. These are described subsequently in section~\ref{sec:context}.

Note that since the encoder is an RNN, ${\bf
  h}_{{\left(b-1\right)W}\cdots bW}$ is actually a function of the
entire input, ${\bf x}_{1 \cdots bW}$ so far.
Correspondingly, ${\bf s}_m$ is a function of the labels emitted so
far, and the entire input seen so far.\footnote{For the
  first output step of a block it includes only the input
  seen until the end of the last block.} Similarly, ${\bf h'}_m$ is a
function of the labels emitted so far and the entire input
seen so far.

\subsection{Computing $f_{context}$\label{sec:context}}
We first describe how the context vector is computed by an attention
model similar to earlier work~\cite{chorowski-nips-2014,bahdanau-iclr-2015,chan2015listen}.
We call this model the MLP-attention model.  

In this model the context vector ${\bf c}_m$ is in computed in two
steps - first a normalized attention vector ${\bf \alpha}_m$ is
computed from the state ${\bf s}_m$ of the transducer and next the
hidden states ${\bf h}_{\left(b-1\right)W+1 \cdots bW}$ of the encoder
for the current block are linearly combined using ${\bf \alpha}$ and
used as the context vector.  To compute ${\bf \alpha}_m$, a
multi-layer perceptron computes a scalar value, $e^m_j$ for each pair
of transducer state ${\bf s}_m$ and encoder ${\bf h}_{(b-1)W+j}$.  The
attention vector is computed from the scalar values, $e^m_j$,
$j=1\cdots W$.  Formally:
\begin{eqnarray}
e^m_j = f_{attention}\left({\bf s}_m, {\bf h}_{\left(b-1\right)W+j};{\bf \theta}\right)\\
{\bf \alpha}_m = softmax\left(\left[e^m_1; e^m_2; \cdots e^m_W\right]\right) \\
{\bf c}_m = \sum_{j=1}^W \alpha^m_j {\bf h}_{\left(b-1\right)W+j}
\end{eqnarray}

We also experimented with using a simpler model for $f_{attention}$ that
computed $e^m_j={\bf s}_m^T {\bf h}_{\left(b-1\right)W+j}$. We refer to this model
as DOT-attention model.

Both of these attention models have two shortcomings. Firstly there is
no explicit mechanism that requires the attention model to move its
focus forward, from one output time step to the next.  Secondly, the
energies computed as inputs to the softmax function, for different
input frames $j$ are independent of each other at each time step, and
thus cannot modulate (e.g., enhance or suppress) each other, other
than through the softmax function. Chorowski et. al.~\cite{chorowski-nips-2015}
ameliorate the second problem by using a convolutional operator
that affects the attention at one time step using the attention at
the last time step.

We attempt to address these two shortcomings using a new attention
mechanism.  In this model, instead of feeding $\left[e^m_1; e^m_2;
  \cdots e^m_W\right]$ into a softmax, we feed them into a recurrent
neural network with one hidden layer that outputs the softmax
attention vector at each time step. Thus the model should be able to
modulate the attention vector both within a time step and across time
steps. This attention model is thus more general than the convolutional
operator of Chorowski et. al. (2015), but it can only be applied to the case
where the context window size is constant. We refer to this model as
LSTM-attention.

\subsection{Addressing End of Blocks}
Since the model only produces a small sequence of output tokens in
each block, we have to address the mechanism for shifting the
transducer from one block to the next. We experimented with three
distinct ways of doing this. In the first approach, we introduced
no explicit mechanism for end-of-blocks, hoping that the 
transducer neural network would implicitly learn a model from the
training data. In the second approach we added end-of-block symbols,
\eob, to the label sequence to demarcate the end of blocks, and we
added this symbol to the target dictionary.  Thus the softmax function
in Equation~\ref{f_softmax} implicitly learns to either emit a token,
or to move the transducer forward to the next block. In the third approach,
we model moving the transducer forward, using a separate logistic function of the
attention vector. The target of the logistic function is 0 or 1
depending on whether the current step is the last step in the block or
not.

\subsection{Training\label{noalignment}}
In this section we show how the Neural Transducer model can
be trained.

The probability of the output sequence $\tilde{y}_{1..S}$, given
${\bf x}_{1 \cdots L}$ is
as follows\footnote{Note that this equation implicitly incorporates the prior for
alignments within the equation}:
\begin{equation}\label{eqn:noalign}
p\left(\tilde{y}_{1 \cdots S} | {\bf x}_{1 \cdots L}\right) = \sum_{y \in \mathcal{Y}} p\left(y_{1 \cdots \left(S+B)\right)} | {\bf x}_{1 \cdots L} \right)
\end{equation}
In theory, we can train the model by maximizing the log of equation~\ref{eqn:noalign}.
The gradient for the log likelihood can easily be expressed as follows:
\begin{equation}\label{eqn:update}
\frac{d}{d \theta} \log {{p\left(\tilde{y}_{1 \cdots S} | {\bf x}_{1 \cdots L}\right)}} = 
\sum_{y \in \mathcal{Y}} p\left(y_{1 \cdots \left(S+B)\right)} | {\bf x}_{1 \cdots L}, \tilde{y}_{1 \cdots S} \right) \frac{d}{d \theta} \log {{p\left(y_{1 \cdots (S+B)} | {\bf x}_{1 \cdots L}\right)}}
\end{equation}
Each of the latter term in the sum on the right hand side can be computed, by backpropagation,
using $y$ as the target of the model.  However, the marginalization is intractable because
of the sum over a combinatorial number of alignments. Alternatively, the gradient can
be approximated by sampling from the posterior distribution (i.e.
$p\left(y_{1 \cdots \left(S+B)\right)} | {\bf x}_{1 \cdots L}, \tilde{y}_{1 \cdots S} \right)$).
However, we found this had very large noise in the learning and the gradients were often
too biased, leading to the models that rarely achieved decent accuracy.

Instead, we attempted to maximize the probability in equation~\ref{eqn:noalign} by
computing the sum over only one term - corresponding to the $y_{1 \cdots S}$ with the
highest posterior probability. Unfortunately, even doing this exactly is computationally
infeasible because the number of possible alignments is combinatorially large
and the problem of finding the best alignment cannot be decomposed to easier
subproblems. So we use an algorithm that finds the approximate best alignment
with a dynamic programming-like algorithm that we describe in the next paragraph.

At each block, $b$, for each output position $j$, this algorithm keeps track of
the approximate best hypothesis $h(j,b)$ that represents the best partial alignment
of the input sequence $\tilde{y}_{1 \cdots j}$ to the partial input ${\bf x}_{1 \cdots bW}$.
Each hypothesis, keeps track of the best alignment $y_{1 \cdots (j+ b)}$ that it represents,
and the recurrent states of the decoder at the last time step, corresponding to this alignment.
At block $b+1$, all hypotheses $h(j,b), j <=\min\left(b\left(M-1 \right),S\right)$ are extended by
at most $M$ tokens using their recurrent states, to compute
$h(j,b+1), h(j+1, b+1) \cdots h(j+M,b+1)$\footnote{Note the minutiae that each of
these extensions ends with $\eob$ symbol.}. For each position $j', j' <=\min\left((b+1)\left(M-1 \right),S\right)$
the highest log probability hypothesis $h(j',b+1)$ is kept\footnote{We
also experimented with sampling from the extensions in proportion to the probabilities,
but this did not always improve results.}.  The alignment from the best hypothesis
$h(S,B)$ at the last block is used for training.


In theory, we need to compute the alignment for each sequence when it is trained,
using the model parameters at that time.  In practice, we batch the alignment inference
steps, using parallel tasks, and cache these alignments. Thus alignments are computed
less frequently than the model updates - typically every 100-300 sequences. This
procedure has the flavor of experience replay from Deep Reinforcement learning work~\cite{dqn}.

\subsection{Inference}
For inference, given the input acoustics ${\bf x}_{1 \cdots L}$, and
the model parameters, ${\bf \theta}$, we find the sequence of labels
$y_{1..M}$ that maximizes the probability of the labels, conditioned
on the data, i.e.,
\begin{equation}
\tilde{y}_{1 \cdots S}= \arg\max_{y_{1 \cdots S'}, e_{1 \cdots N}}
   \sum_{b=1}^N \log{p\left(y_{e_{\left(b-1\right)+1} \cdots e_b}|
          {\bf x}_{1 \cdots bW}, y_{1 \cdots e_{\left(b-1\right)}} \right)}
\end{equation}
Exact inference in this scheme is computationally expensive because 
the expression for log probability does not permit decomposition into
smaller terms that can be independently computed. Instead, each
candidate, $y_{1 \cdot S'}$, would have to be tested independently,
and the best sequence over an exponentially large number of sequences
would have to be discovered. Hence, we use a beam search heuristic
to find the ``best'' set of candidates. To do this, at each output
step $m$, we keep a heap of alternative $n$ best prefixes, and extend
each one by one symbol, trying out all the possible alternative
extensions, keeping only the best $n$ extensions. Included in the
beam search is the act of moving the attention to the next input block.
The beam search ends either when the sequence is longer than 
a pre-specified threshold, or when the end of token symbol is produced
at the last block.

\section{Experiments and Results}
\subsection{Addition Toy Task}
We experimented with the Neural Transducer on the toy task of adding two
three-digit decimal numbers. The second number is presented in the reverse
order, and so is the target output. Thus the model can produce the first output as
soon as the first digit of the second number is observed. The model is able
to learn this task with a very small number of units (both encoder and transducer
are 1 layer unidirectional LSTM RNNs with 100 units).

As can be seen below, the model learns to output the digits as soon as the
required information is available. Occasionally the model waits an extra
step to output its target symbol. We show results (blue) for four different
examples (red). A block window size of W=1 was used, with M=8.

\begin{table}[h!]
\begin{center}
\begin{footnotesize}
\begin{tabular}{cccccc||cccccccc}
\hline
		\color{red}2 & \color{red} + & \color{red} 7 & \color{red}  2    & \color{red}5   & \color{red} <s> & \color{red}  2 & \color{red} 2 & \color{red} 7 & \color{red} + & \color{red} 3 & \color{red} <s>  & \color{red} \\
	\color{blue}<e>& \color{blue}<e>& \color{blue}<e>& \color{blue}  9<e> & \color{blue}2<e>& \color{blue} 5<e> & \color{blue} <e>& \color{blue}<e>& \color{blue}<e>& \color{blue}<e>& \color{blue}<e>& \color{blue}032<e>& \color{blue} \\
\hline
  \color{red}1 & \color{red} 7 & \color{red} 4 & \color{red} + & \color{red} 3 & \color{red} <s>  & \color{red}   4 & \color{red} 0 & \color{red} + & \color{red} 2 & \color{red} 6  & \color{red}  2 & \color{red} <s> \\
	\color{blue}<e>& \color{blue}<e>& \color{blue}<e>& \color{blue}<e>& \color{blue}<e>& \color{blue}771<e>& \color{blue}  <e>& \color{blue}<e>& \color{blue}<e>& \color{blue}<e>& \color{blue}2<e>& \color{blue}0<e>& \color{blue}3<e> \\
\hline
\end{tabular}
\end{footnotesize}
\end{center}
\end{table}
The model achieves an error rate of 0\% after 500K examples.

\subsection{TIMIT}
\label{eperiments}
We used TIMIT, a standard benchmark for speech recognition, for our larger
experiments.
Log Mel filterbanks were computed every 10ms as inputs
to the system. The targets were the 60 phones defined for the TIMIT dataset
({\it h\#} were relabelled as {\it pau}).

We used stochastic gradient descent with momentum with a batch size of
one utterance per training step. An initial learning rate of 0.05, and
momentum of 0.9 was used. The learning rate was reduced by a factor of 0.5
every time the average log prob over the validation set decreased
\footnote{Note the TIMIT provides a validation set, called the {\it dev} set.
We use these terms interchangeably.}.  The decrease was applied for a
maximum of 4 times. The models were trained for 50 epochs and the
parameters from the epochs with the best dev set log prob were used for decoding.

We trained a Neural Transducer with three layer LSTM RNN coupled
to a three LSTM layer unidirectional encoder RNN, and achieved a PER of 20.8\%
on the TIMIT test set. This model used the LSTM attention mechanism. Alignments
were generated from a model that was updated after every 300 steps of
Momentum updates.  Interestingly, the alignments generated by the model are
very similar to the alignments produced by a Gaussian Mixture Model-Hidden Markov
Model (GMM-HMM) system that we trained using the Kaldi toolkit -- even though
the model was trained entirely discriminatively. The small differences in
alignment correspond to an occasional phoneme emitted slightly later by our
model, compared to the GMM-HMM system.

We also trained models using alignments generated from the
GMM-HMM model trained on Kaldi.  The frame level alignments from Kaldi
were converted into block level alignments by assigning each phone in
the sequence to the block it was last observed in. The same architecture
model described above achieved an accuracy of 19.8\% with these alignments.

We did further experiments to assess the properties of the model.  In order
to avoid the computation associated with finding the best alignments, we ran
these experiments using the GMM-HMM alignments.

Table~\ref{tab:block_recurrence} shows a comparison of our method
against a basic implementation of a sequence-to-sequence model that
produces outputs for each block independent of the other blocks,
and concatenates the produced sequences. Here, the sequence-to-sequence
model produces the output conditioned on the state of the encoder
at the end of the block.  Both models used an encoder
with two layers of 250 LSTM cells, without attention. The standard
sequence-to-sequence model performs significantly worse than our
model -- the recurrent connections of the transducer across blocks are
clearly helpful in improving the accuracy of the model.

\begin{table}[h!]
\caption{\small Impact of maintaining recurrent state of transducer across
  blocks on the PER.  This table shows that maintaining the state of
  the transducer across blocks leads to much better results. For this
  experiment, a block size (W) of 15 frames was used. The reported number
  is the median of three different runs.\label{tab:block_recurrence}}
\begin{center}
\begin{tabular}{ccc}
{\bf W} & {\bf BLOCK-RECURRENCE} & {\bf PER}\\
\hline
15 & No & 34.3\\
15 & Yes & 20.6\\\hline
\end{tabular}
\end{center}
\end{table}

\begin{figure}[h!]
\begin{center}
 \includegraphics[width=0.6\textwidth]{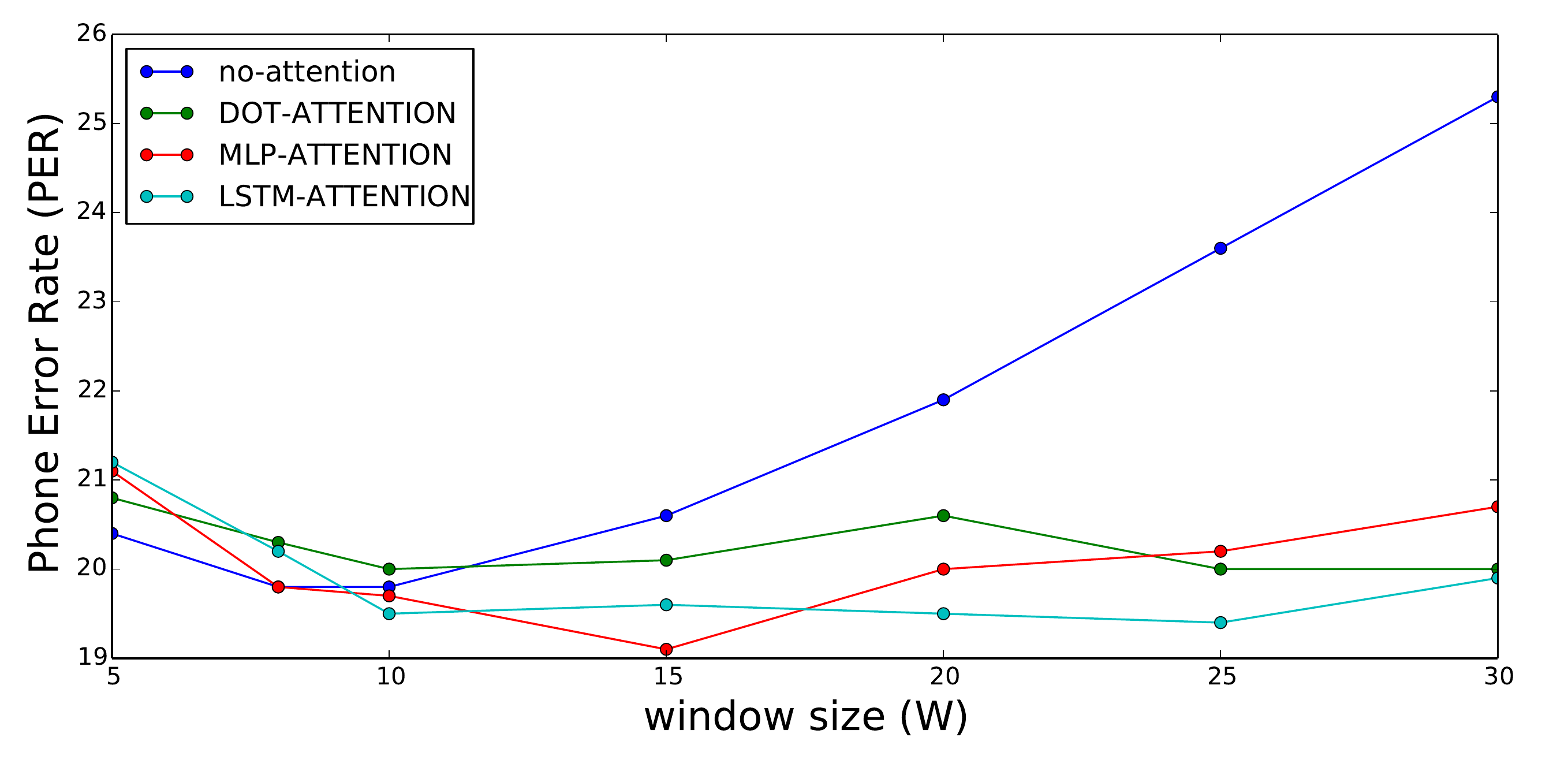}\\
\end{center}
\caption{\small Impact of the number of frames (W) in a block and attention mechanism on PER.
  Figure shows the PER on the validation set as function of W, with and without the
  use of different types of attention mechanisms.  Each number is the
  median value from three experiments.\label{fig:block_size}}
\end{figure}

Figure~\ref{fig:block_size} shows the impact of block size on the
accuracy of the different transducer variants that we used.
See Section~\ref{sec:context} for a description of the \{DOT,MLP,LSTM\}-attention models.
All models used a two LSTM layer encoder and a two LSTM layer transducer.
The model is sensitive to the choice of the block size, when no attention
is used. However, it can be seen that with an appropriate choice of
window size (W=8), the Neural Transducer without attention can match the accuracy
of the attention based Neural Transducers. Further exploration of this
configuration should lead to improved results.

When attention is used in the transducer, the precise value of the block size becomes less
important. The LSTM-based attention model seems to be more consistent compared to the other
attention mechanisms we explored. Since this model performed best with W=25, we used this
configuration for subsequent experiments.

Table~\ref{tab:architecture} explores the impact of the number of layers
in the transducer and the encoder on the PER. A three layer encoder coupled to
a three layer transducer performs best on average.  Four layer transducers
produced results with higher spread in accuracy -- possibly because of the
more difficult optimization involved.
Thus, the best average PER we achieved (over 3 runs) was {\bf 19.8\%} on the TIMIT
test set. These results could probably be improved with other regularization
techniques, as reported by~\cite{chorowski-nips-2015} but we did not pursue
those avenues in this paper.
\begin{table}[h!]
\caption{\small Impact of architecture on PER. Table shows the PER on the dev set
 as function of the number of layers (2 or 3) in the encoder
 and the number of layers in the transducer (1-4).\label{tab:architecture}}
\begin{center}
\begin{tabular}{ccccc}
	{\bf \# of layers in encoder / transducer} & {\bf 1} & {\bf 2} & {\bf 3} & {\bf 4}\\
\hline 
 {\bf 2 }&  & 19.2 & 18.9 & 18.8 \\
 {\bf 3 }&  & 18.5 & {\bf 18.2} & 19.4 \\
\hline 
\end{tabular}
\end{center}
\end{table}

For a comparison with previously published sequence-to-sequence models
on this task, we used a three layer bidirectional LSTM encoder with
250 LSTM cells in each direction and achieved a PER of 18.7\%. By
contrast, the best reported results using previous
sequence-to-sequence models are 17.6\%~\cite{chorowski-nips-2015}.
However, this result comes with more careful training techniques
than we attempted for this work. Given the high variance in results
from run to run on TIMIT, these numbers are quite promising.

\section{Discussion}
One of the important side-effects of our model using partial conditioning
with a blocked transducer is that it naturally alleviates the problem 
of ``losing attention'' suffered by sequence-to-sequence
models. Because of this, sequence-to-sequence models perform
worse on longer utterances~\cite{chorowski-nips-2015,chan2015listen}.
This problem is automatically tackled in our model because each new block
automatically shifts the attention monotonically forward. Within a block,
the model learns to move attention forward from one step to the next,
and the attention mechanism rarely suffers,
because both the size of a block, and the number of output steps for a
block are relatively small. As a result, error in attention in one block,
has minimal impact on the predictions at subsequent blocks. 

Finally, we note that increasing the block size, $W$, so that it is as
large as the input utterance makes the model similar to vanilla 
end-to-end models~\cite{chorowski-nips-2014,chan2015listen}.

\section{Conclusion}
We have introduced a new model that uses partial conditioning on
inputs to generate output sequences. This allows the model to produce
output as input arrives. This is useful for speech recognition systems and
will also be crucial for future generations of online speech translation
systems. Further it can be useful for performing transduction over long
sequences -- something that is possibly difficult for sequence-to-sequence
models. We applied the model to a toy task of addition, and to a phone
recognition task and showed that is can produce results comparable to
the state of the art from sequence-to-sequence models.

\footnotesize
\bibliography{cites}

\begin{thebibliography}{10}

\bibitem{bahdanau-iclr-2015}
Dzmitry Bahdanau, Kyunghyun Cho, and Yoshua Bengio.
\newblock {Neural Machine Translation by Jointly Learning to Align and
  Translate}.
\newblock In {\em {International Conference on Learning Representations}},
  2015.

\bibitem{BahdanauCSBB15}
Dzmitry Bahdanau, Jan Chorowski, Dmitriy Serdyuk, Philemon Brakel, and Yoshua
  Bengio.
\newblock End-to-end attention-based large vocabulary speech recognition.
\newblock In {\em http://arxiv.org/abs/1508.04395}, 2015.

\bibitem{chan2015listen}
William Chan, Navdeep Jaitly, Quoc~V Le, and Oriol Vinyals.
\newblock Listen, attend and spell.
\newblock {\em arXiv preprint arXiv:1508.01211}, 2015.

\bibitem{cho-emnlp-2014}
Kyunghyun Cho, Bart van Merrienboer, Caglar Gulcehre, Dzmitry Bahdanau, Fethi
  Bougares, Holger Schwen, and Yoshua Bengio.
\newblock {Learning Phrase Representations using RNN Encoder-Decoder for
  Statistical Machine Translation}.
\newblock In {\em {Conference on Empirical Methods in Natural Language
  Processing}}, 2014.

\bibitem{chorowski-nips-2014}
Jan Chorowski, Dzmitry Bahdanau, Kyunghyun Cho, and Yoshua Bengio.
\newblock {End-to-end Continuous Speech Recognition using Attention-based
  Recurrent NN: First Results}.
\newblock In {\em {Neural Information Processing Systems: Workshop Deep
  Learning and Representation Learning Workshop}}, 2014.

\bibitem{chorowski-nips-2015}
Jan Chorowski, Dzmitry Bahdanau, Dmitriy Serdyuk, Kyunghyun Cho, and Yoshua
  Bengio.
\newblock {Attention-Based Models for Speech Recognition}.
\newblock In {\em {Neural Information Processing Systems}}, 2015.

\bibitem{graves2013speech}
Alan Graves, Abdel{-}rahman Mohamed, and Geoffrey Hinton.
\newblock Speech recognition with deep recurrent neural networks.
\newblock In {\em Acoustics, Speech and Signal Processing (ICASSP), 2013 IEEE
  International Conference on}, pages 6645--6649. IEEE, 2013.

\bibitem{graves-icml-2012}
Alex Graves.
\newblock {Sequence Transduction with Recurrent Neural Networks}.
\newblock In {\em {International Conference on Machine Learning: Representation
  Learning Workshop}}, 2012.

\bibitem{graves-icassp-2013}
Alex Graves, Abdel{-}rahman Mohamed, and Geoffrey Hinton.
\newblock {Speech Recognition with Deep Recurrent Neural Networks}.
\newblock In {\em {IEEE International Conference on Acoustics, Speech and
  Signal Processing}}, 2013.

\bibitem{graves2014neural}
Alex Graves, Greg Wayne, and Ivo Danihelka.
\newblock Neural turing machines.
\newblock {\em arXiv preprint arXiv:1410.5401}, 2014.

\bibitem{hinton2012deep}
Geoffrey Hinton, Li~Deng, Dong Yu, George~E. Dahl, Abdel{-}rahman Mohamed,
  Navdeep Jaitly, Andrew Senior, Vincent Vanhoucke, Patrick Nguyen, Tara~N.
  Sainath, and Brian Kingsbury.
\newblock Deep neural networks for acoustic modeling in speech recognition: The
  shared views of four research groups.
\newblock {\em Signal Processing Magazine, IEEE}, 29(6):82--97, 2012.

\bibitem{dqn}
Volodymyr Mnih, Koray Kavukcuoglu, David Silver, Alex Graves, Ioannis
  Antonoglou, Daan Wierstra, and Martin~A. Riedmiller.
\newblock Playing atari with deep reinforcement learning.
\newblock {\em CoRR}, abs/1312.5602, 2013.

\bibitem{neelakantan2015neural}
Arvind Neelakantan, Quoc~V Le, and Ilya Sutskever.
\newblock Neural programmer: Inducing latent programs with gradient descent.
\newblock {\em arXiv preprint arXiv:1511.04834}, 2015.

\bibitem{reed2015neural}
Scott Reed and Nando de~Freitas.
\newblock Neural programmer-interpreters.
\newblock {\em arXiv preprint arXiv:1511.06279}, 2015.

\bibitem{sordoni2015neural}
Alessandro Sordoni, Michel Galley, Michael Auli, Chris Brockett, Yangfeng Ji,
  Margaret Mitchell, Jian-Yun Nie, Jianfeng Gao, and Bill Dolan.
\newblock A neural network approach to context-sensitive generation of
  conversational responses.
\newblock {\em arXiv preprint arXiv:1506.06714}, 2015.

\bibitem{sukhbaatar2015end}
Sainbayar Sukhbaatar, Jason Weston, Rob Fergus, et~al.
\newblock End-to-end memory networks.
\newblock In {\em Advances in Neural Information Processing Systems}, pages
  2431--2439, 2015.

\bibitem{sutskever-nips-2014}
Ilya Sutskever, Oriol Vinyals, and Quoc~V. Le.
\newblock {Sequence to Sequence Learning with Neural Networks}.
\newblock In {\em {Neural Information Processing Systems}}, 2014.

\bibitem{vinyals2015grammar}
Oriol Vinyals, Lukasz Kaiser, Terry Koo, Slav Petrov, Ilya Sutskever, and
  Geoffrey Hinton.
\newblock Grammar as a foreign language.
\newblock In {\em NIPS}, 2015.

\bibitem{vinyals2015neural}
Oriol Vinyals and Quoc~V. Le.
\newblock A neural conversational model.
\newblock In {\em ICML Deep Learning Workshop}, 2015.

\bibitem{vinyals-arvix-2014}
Oriol Vinyals, Alexander Toshev, Samy Bengio, and Dumitru Erhan.
\newblock {Show and Tell: A Neural Image Caption Generator}.
\newblock In {\em {IEEE Conference on Computer Vision and Pattern
  Recognition}}, 2015.

\bibitem{zaremba2015reinforcement}
Wojciech Zaremba and Ilya Sutskever.
\newblock Reinforcement learning neural turing machines.
\newblock {\em arXiv preprint arXiv:1505.00521}, 2015.

\end{thebibliography}
\bibliographystyle{plain}

\end{document}